\documentclass[11pt]{article}

\usepackage[final]{acl}

\usepackage{times}
\usepackage{latexsym}

\usepackage[T1]{fontenc}

\usepackage[utf8]{inputenc}

\usepackage{microtype}

\usepackage{inconsolata}

\usepackage{graphicx}

\usepackage{amsmath,amssymb}
\usepackage{xspace}
\usepackage{tikz}
\usetikzlibrary{arrows.meta,positioning}
\usepackage{listings}
\usepackage{booktabs,tabularx,array}
\usepackage[table]{xcolor}
\usepackage{arydshln}

\lstdefinestyle{prompt}{
    basicstyle=\ttfamily\footnotesize,
    columns=fullflexible,
    keepspaces=true,
    breaklines=true,
    breakautoindent=false,
    breakindent=0pt,
    showstringspaces=false,
    frame=single,
    framerule=0.3pt,
    rulecolor=\color{black!25},
    backgroundcolor=\color{black!2},
    framesep=4pt,
    aboveskip=5pt,
    belowskip=6pt
}

\newcommand{\method}{SheetAlign\xspace}
\newcommand{\sbench}{\texttt{SpreadsheetBench}}
\newcommand{\dbench}{\texttt{Domain-Spreadsheet}}


\title{SheetAlign: Grounded Interaction and Stage-Isolated Learning for Spreadsheet Agents}

\author{
 {
  \textbf{Yuxiang Wang}, 
  \textbf{Junhao Gan}, 
  \textbf{Jianzhong Qi} 
  }
  \\
  {The University of Melbourne} 
  \\ \texttt{yuxiang.wang8@student.unimelb.edu.au}\\ 
   \texttt{\{junhao.gan, jianzhong.qi\}@unimelb.edu.au}
}

\begin{document}
\maketitle
\begin{abstract}
Spreadsheets are widely used for data analysis and supporting decision-making. Automating spreadsheet analysis while protecting sensitive data and controlling deployment costs motivates locally deployed open-weight large language models (LLMs). Meanwhile, complex spreadsheet tasks require both task reasoning and workbook understanding, while predefined operation interfaces impose additional tool-selection and invocation demands. 
We introduce \method, a spreadsheet manipulation framework with stage-isolated policy learning, organized around \emph{where} to operate, \emph{what} to change, and \emph{how} to implement transformations. Task-guided inspection and API grounding anchor code generation in workbook observations and available operations, while stage-isolated learning optimizes program generation and review using edit-progress rewards. At inference, execution consistency selects programs through agreement among workbook states. On SpreadsheetBench, our grounded task interface improves Qwen3-4B's accuracy from $13.5\%$ to $29.2\%$ on average without training. With learning and execution consistency, \method achieves $32.8\%$ accuracy on average on SpreadsheetBench and $30.1\%$ on Domain-Spreadsheet, outperforming comparable baselines. The code will be made publicly available upon publication.
\end{abstract}

\section{Introduction}
\label{sec:introduction}

Spreadsheets support data analysis and operational workflows across many domains. LLM agents allow users to perform these tasks through natural-language instructions without manually constructing formulas or scripts \citep{sheetcopilot,sheetagent}. Keeping sensitive workbooks local under limited computing budgets motivates compact, open-weight LLMs.

\begin{figure}[t]
    \centering
    \includegraphics[width=0.80\columnwidth]{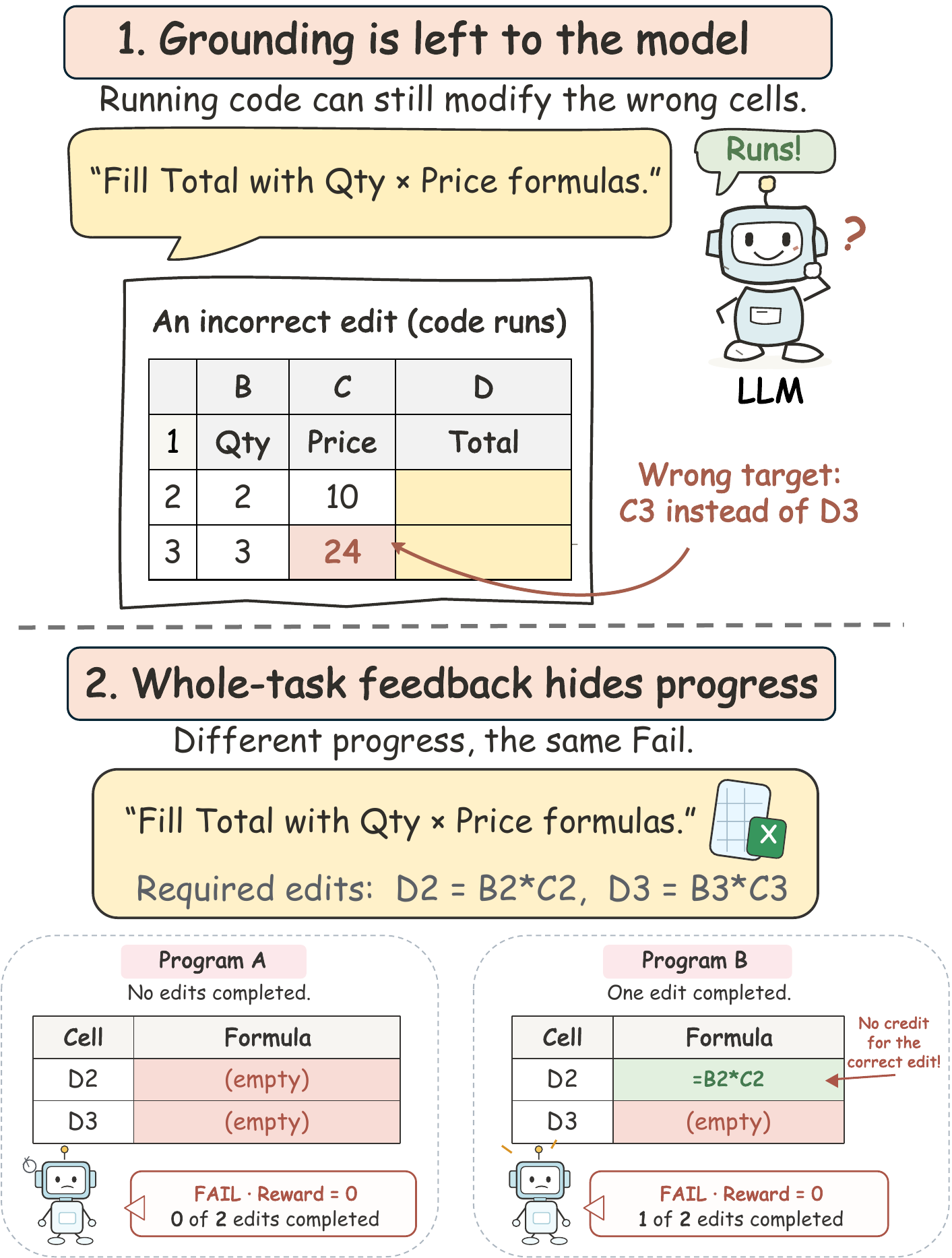}
    \caption{Motivations for \method. Upper: Executable programs may modify unintended cells without task-to-edit grounding. Lower: Whole-task rewards overlook partially completed edits. }
    \label{fig:intro}
\end{figure}

\textbf{Challenge 1: Task-to-Edit Grounding.}
Real-world spreadsheet requests often include previous attempts, encountered problems, and examples of desired outputs \citep{spreadsheetbench}. Existing agents rely on predefined actions or program execution with iterative verification \citep{sheetcopilot,sheetagent,sheetbrain}. Regardless of the interface, agents must identify relevant workbook regions, determine required edits, and map them to appropriate operations. Resolving these interdependent decisions within complex workbooks is particularly challenging for compact models with limited reasoning capabilities. As illustrated in Figure~\ref{fig:intro} (upper part), a program may execute successfully yet modify unintended cells. This motivates grounding code generation in focused workbook observations and relevant API references, guiding \emph{where} and \emph{what} to edit while preserving implementation flexibility.

\textbf{Challenge 2: Limited Training Feedback.}
Spreadsheet tasks often require multiple coordinated edits, and partially correct transformations may still fail the overall task. As shown in Figure~\ref{fig:intro} (lower part), outcome-based rewards cannot distinguish different degrees of partial completion, providing limited guidance on remaining edits \citep{spreadsheetrl}. Moreover, program generation and review involve distinct decisions: generation constructs the requested transformation through programs, while program review identifies remaining intent misalignment for generated programs without disrupting correct edits. These observations motivate our edit-progress rewards and stage-isolated learning, as detailed in Section~\ref{sec:method}.

\textbf{Challenge 3: Generation Instability.}
Due to the stochastic nature of LLMs, repeated generations for the same spreadsheet task may yield both correct and incorrect transformations \citep{selfconsist}. This variability makes individual predictions unreliable, motivating execution consistency to improve accuracy and reliability through agreement among resulting workbooks.

We introduce \method, a spreadsheet manipulation framework combining a grounded task interface, stage-isolated policy learning, and execution consistency. It organizes each task around \emph{where} to operate, \emph{what} to change, and \emph{how} to implement the transformation. Task-guided inspection and API grounding guide code generation with focused workbook observations and relevant API references, while execution feedback supports diagnostic repair. We separately train a generation policy to construct transformations and a review policy to refine executable programs, both using edit-progress rewards. At inference, execution consistency selects a program through agreement among resulting workbooks, followed by a review pass to address intent misalignment.

Our contributions are summarized as follows:
\textbf{(1) Grounded task interface.}
We introduce an interface that connects task-specific edit requirements with focused workbook observations and relevant APIs to guide code generation.
\textbf{(2) Stage-isolated policy learning.}
We separately optimize generation and review policies using edit-progress rewards to recognize partial completion and address remaining intent misalignment.
\textbf{(3) Execution consistency and empirical validation.}
We leverage agreement among resulting workbooks to mitigate generation instability and demonstrate its effectiveness on real-world spreadsheet benchmarks.

\section{Related Work}

\paragraph{Spreadsheet Agents.}
Existing spreadsheet agents employ different approaches to spreadsheet manipulation. InstructExcel generates Office Scripts from natural-language instructions \citep{instructexcel}, while SheetCopilot plans over predefined atomic actions \citep{sheetcopilot}. SheetAgent extends these capabilities through planning, information gathering, and program execution \citep{sheetagent}. Other work emphasizes workbook understanding. SpreadsheetLLM develops compact representations of large and irregular spreadsheets \citep{spreadsheetllm}, while SheetBrain integrates query-aware workbook understanding with Python execution and validation \citep{sheetbrain}. Recent benchmarks further cover real-world manipulation, business workflows, financial modeling, and workbook artifact creation \citep{spreadsheetbench,spreadsheetv2,bluefin,wtmbench}. Our work instead emphasizes explicit task-to-edit grounding, identifying what to read, where to write, and which operations to perform before code generation.

\paragraph{Spreadsheet Action Interfaces.}
At the execution level, spreadsheet agents use predefined actions or executable programs. Predefined actions provide structured operations, while program-based interfaces support compositional transformations \citep{sheetcopilot,instructexcel,sheetagent}. Spreadsheet-RL combines spreadsheet-native tools with sandboxed Python execution in a multi-turn environment \citep{spreadsheetrl}. Beyond spreadsheets, CodeAct demonstrates executable code as a general agent action interface \citep{codeact}. However, action interfaces alone cannot specify how task-specific edits should be implemented in the current workbook. To address this, \method grounds identified edits in focused workbook observations and relevant API references to guide program generation.

\begin{figure*}[t]
    \centering
    \includegraphics[width=0.85\textwidth]{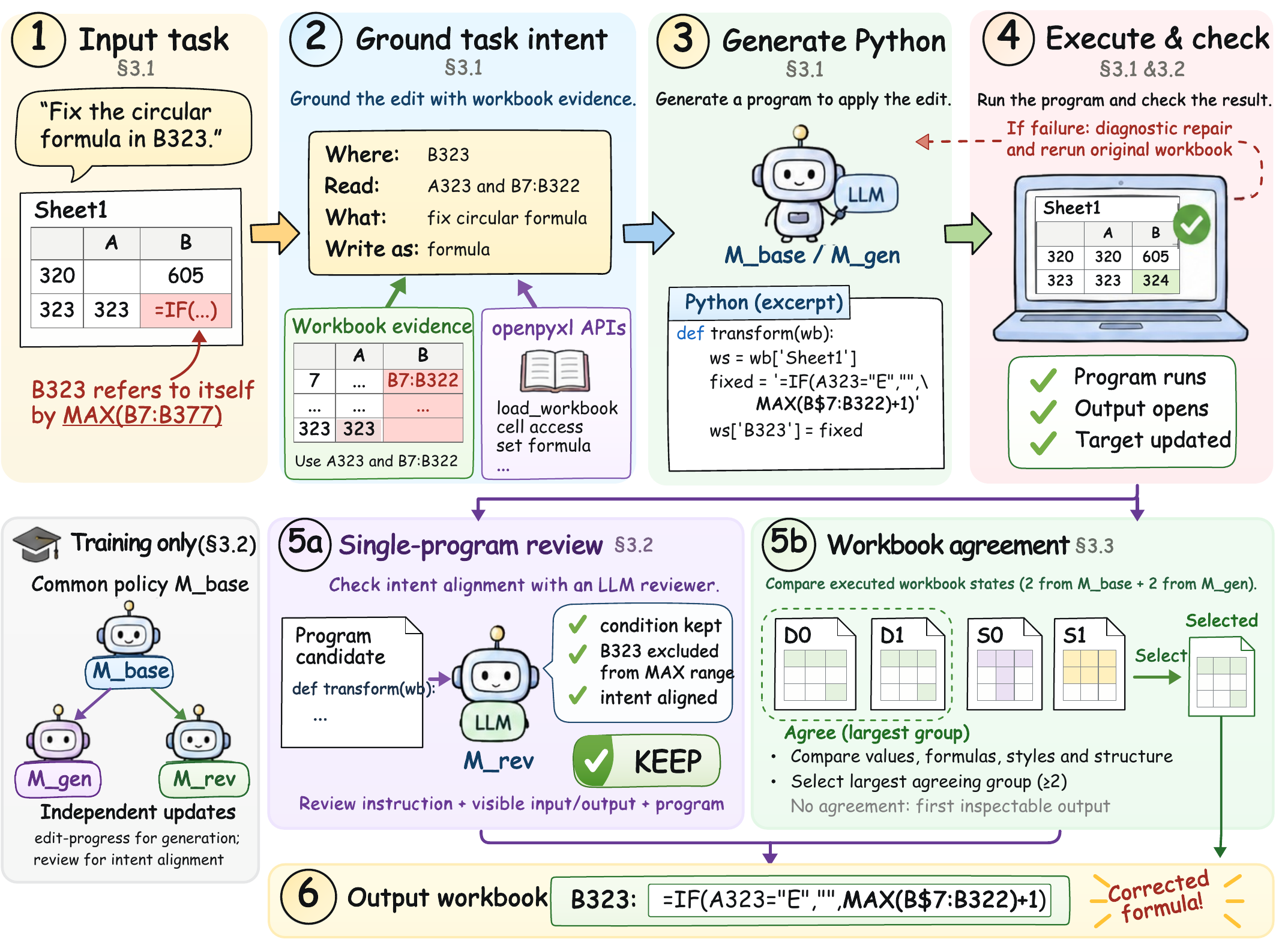}
    \caption{Overview of \method, integrating task grounding, stage-isolated learning, and execution consistency.}
    \label{fig:method-overview}
\end{figure*}

\paragraph{Learning for Spreadsheet Generation and Revision.}
Execution-based learning has been explored for program generation, including spreadsheet transformations. CodeRL uses functional correctness feedback \citep{coderl}, while Spreadsheet-RL trains spreadsheet agents using execution and workbook-level outcomes \citep{spreadsheetrl}. Such feedback provides limited guidance on which required workbook edits have been completed. For revision, Self-Debugging refines programs through execution feedback and explanations \citep{selfdebug}, while SCoRe trains language models for self-correction \citep{SCoRe}. Building on these approaches, \method combines diagnostic repair for execution failures with a dedicated review policy for intent misalignment in executable programs. Stage-isolated learning further optimizes generation and review separately using edit-progress rewards.

\paragraph{Execution-Based Candidate Selection.}
Prior work has explored consistency across multiple candidates to improve reliability. Self-consistency selects answers through agreement across sampled reasoning paths \citep{selfconsist}, while other approaches compare execution outputs or train candidate selectors \citep{CSCSQL,tablap,dpc}. Spreadsheet programs may modify cell values, formulas, and structure, making agreement over the resulting workbooks more informative than answer-level agreement. Accordingly, \method selects programs by comparing their resulting workbooks.

\section{Method}
\label{sec:method}

Given a natural-language instruction $x$, an input workbook $W$, and task-provided target locations $T$, our goal is to produce a transformed workbook $W'$ satisfying the user's requirements. The target locations specify where results are expected, such as a worksheet cell or range, while the requested changes may involve values, formulas, formatting, or workbook structure. We introduce \method (shown in Figure~\ref{fig:method-overview}), a spreadsheet agent built around three components: a grounded task interface, stage-isolated policy learning, and execution consistency selection.

The model expresses each transformation as Python code using \texttt{openpyxl}, while a shared harness provides task-relevant workbook observations and API references, executes generated programs, and returns diagnostics. Within this harness, we study two separately trained policies: a generation policy that produces spreadsheet transformations and a program review policy that revisits generated programs. The policies are trained independently, allowing generation and review to be studied separately. At inference, we select among multiple complete candidates by agreement between their resulting workbook states.

\subsection{Grounded Task Interface}
\label{sec:grounding}

\paragraph{Task-guided workbook inspection.}
Before generating code, the model summarizes the task requirements
and relevant workbook information using a compact template:

\begin{center}
\small
\begin{tabular}{@{}ll@{}}
\textbf{Where} & target region to modify \\
\textbf{From} & workbook context to inspect \\
\textbf{What} & requested transformation \\
\textbf{Write as} & values, formulas, formatting, or structure \\
\textbf{API} & relevant \texttt{openpyxl} functionality \\
\end{tabular}
\end{center}

The template guides workbook inspection and API retrieval for subsequent code generation. For example, if the instruction asks to compute values from two source columns and write it to a target column, the description identifies the source columns, the target region, and the required computation. The harness then provides focused workbook observations, such as nearby headers, cells, formulas, and merged regions. These observations ground code generation in the current workbook state. The full structured representation and observation procedure are given in Appendix~\ref{app:method-interface}.

\paragraph{API grounding.}
The same task summarization also identifies the APIs needed for the task. The harness retrieves a small set of relevant \texttt{openpyxl} API references and provides them to the model together with the focused workbook observations. The model thus receives both the workbook context needed to understand the task and the API guidance needed to implement the transformation, while retaining the flexibility of code generation.


\paragraph{Execution and diagnostic repair.}
The harness executes each generated program and checks for execution failures or invalid workbook outputs. When an issue is detected, a concise diagnostic describing the problem, together with the current program and relevant workbook context, is provided to the \emph{generation policy} introduced later. The same policy revises the program, which is rerun from the original workbook, with up to two repair attempts by default. This failure-driven repair differs from the separately trained \emph{review policy}, also introduced later, which targets intent misalignment in executable programs through a single review pass. Further details are provided in Appendix~\ref{app:diagnositic_repair}.

\subsection{Stage-Isolated Policy Learning}
\label{sec:learning}

We train two separate policies from the same backbone LLM. The
\emph{generation policy} produces spreadsheet programs, while the \emph{review policy}
examines an executable program and decides whether to keep or revise it.
We first obtain a common starting policy and then train the two policies
independently. Each policy is optimized for its own role.

\paragraph{Shared optimization.}
We optimize each trainable policy with group relative policy optimization
(GRPO)~\cite{deepseekmath} and use LoRA
\citep{lora} for parameter-efficient updates.
For each prompt, GRPO samples $K$ (default by 4) responses and compares their rewards within the group. If response $i$ receives reward $r_i$, its group-relative advantage is
\vspace{-0.7cm}
\begin{center}
\small
\begin{equation}
A_i =
\frac{r_i-\bar{r}}{\sigma_r+\epsilon},
\end{equation}
\end{center}

where $\bar{r}$ and $\sigma_r$ are the mean and standard deviation of rewards
for the same prompt. A positive $A_i$ indicates that the response performs
better than the group average.

Let $\rho_{i,t}$ denote the token-level probability ratio between the updated
policy and the policy used to sample response $i$. We define its clipped form as

\vspace{-0.6cm}
\begin{center}
\small
\begin{equation}
\tilde{\rho}_{i,t}
=
\operatorname{clip}
\left(\rho_{i,t},\,1-\epsilon_c,\,1+\epsilon_c\right),
\end{equation}
\end{center}

where $\epsilon_c$ controls the allowed change in the policy. We then optimize

\vspace{-0.6cm}
\begin{center}
\small
\begin{equation}
\mathcal{J}_{\mathrm{GRPO}}
=
\mathbb{E}_{i,t}
\left[
\min\left(
\rho_{i,t}A_i,\,
\tilde{\rho}_{i,t}A_i
\right)
\right].
\end{equation}
\end{center}

The clipped objective limits overly large policy updates while favoring
responses with higher relative rewards. Full optimization details and training settings are provided in Appendix~\ref{app:method-training}.

\paragraph{Common starting policy.}
We first train a common starting policy $M_{base}$ with a diagnostic reward based on task
correctness $R_{\mathrm{correct}}$, execution success $R_{\mathrm{exec}}$, and
output quality $R_{\mathrm{quality}}$, with an additional penalty
$R_{\mathrm{fail}}$ for failed programs. We summarize the reward as

\vspace{-0.7cm}
\begin{center}
\small
\begin{equation}
R_{\mathrm{init}}
=
\alpha R_{\mathrm{correct}}
+\beta R_{\mathrm{exec}}
+\gamma R_{\mathrm{quality}}
-R_{\mathrm{fail}},
\label{eq:initial-reward}
\end{equation}
\end{center}

where $R_{\mathrm{fail}}$ penalizes execution failures according to their type,
and $\alpha$, $\beta$, and $\gamma$ are fixed coefficients. The implementation applies these terms differently depending on the program outcome. This policy provides the common starting point for the generation policy and review policy branches. Appendix~\ref{app:method-reward} gives the exact piecewise reward and coefficients.

\paragraph{Generation policy training.}
The $M_{base}$ learns broad task correctness and program execution,
while the generation policy $M_{gen}$ training focuses on tasks whose solutions remain unstable. For each training task, we sample multiple programs and prioritize cases that contain both a correct program and an executable but incorrect one. These cases show that the $M_{base}$ can solve the task, but does not yet produce the intended result consistently. We also sample coverage tasks across operation categories such as formulas, formatting, lookup, and structural edits to maintain task diversity.

\paragraph{Edit-level reward.}
Whole-task correctness overlooks partially completed solutions.
We evaluate progress over required target-cell edits.
Let $N_r$ be the number of required edits, $N_m$ the number of edits made by
the program, and $N_c$ the number of required edits completed correctly.
We compute

\vspace{-0.7cm}
\begin{center}
\small
\begin{equation}
\mathrm{Precision}=\frac{N_c}{N_m},
\qquad
\mathrm{Recall}=\frac{N_c}{N_r},
\end{equation}
\end{center}

and use their harmonic mean as the edit-level progress reward

\vspace{-0.7cm}
\begin{center}
\small
\begin{equation}
F_1=
\frac{2\,\mathrm{Precision}\,\mathrm{Recall}}
{\mathrm{Precision}+\mathrm{Recall}},
\end{equation}
\end{center}

Precision discourages unnecessary edits, while recall measures completed required edits. Let $H\in\{0,1\}$ indicate task correctness. We then combine $F_1$ and $H$ as

\vspace{-0.7cm}
\begin{center}
\small
\begin{equation}
R_{\mathrm{progress}} =\lambda F_1+(1-\lambda)H,
\label{eq:step-reward}
\end{equation}
\end{center}

where $\lambda=0.75$. During generation and review policy training, $R_{\mathrm{progress}}$ replaces the correctness term $R_{\mathrm{correct}}$ in the initial reward, while the execution and output quality terms remain unchanged.

\paragraph{Review policy training.}
Diagnostic repair handles clear implementation or execution failures, but an executable program may still perform the wrong transformation. We therefore train a separate review policy $M_{rev}$ as a complementary correction stage. Given the user instruction, an executable program, and focused observations of the input and output workbooks covering task-relevant cells, headers, and formulas, the $M_{rev}$ decides whether to keep the program or revise it.

Both the $M_{gen}$ and $M_{rev}$ are continued independently from the same common starting policy $M_{base}$. The $M_{gen}$ learns to produce better initial solutions, while the $M_{rev}$ learns to correct existing executable solutions. Their training is separate, so updates to one do not affect the other.

For review training, we construct training examples from fixed executable
programs. The $M_{rev}$ either keeps the original program or generates a revised
one. The resulting program is executed and evaluated with the same reward function used for generation training, since both branches optimize the correctness of the final workbook. Here, $R_{\mathrm{progress}}$ still serves as the correctness term.

\subsection{Execution Consistency Selection}
\label{sec:consistency}

We further improve inference reliability by selecting among multiple candidate
programs based on their resulting workbooks. We sample two candidates from the
$M_{base}$ and two from the $M_{gen}$. Each candidate undergoes the same grounding, execution, and diagnostic repair procedure.

\paragraph{Workbook agreement.}
For each output workbook that the harness can open and inspect, we construct a deterministic fingerprint from worksheet structure, cell locations, stored values and formulas, and formatting information. Two candidates agree when their fingerprints match, even if their generated programs differ. We select the first candidate from the largest agreement group containing at least two candidates. If no candidates agree, we select the first candidate with a valid output, or the first candidate if none is valid.  After consistency selection, the review policy retains or revises the selected program to address intent misalignment. Further details are provided in Appendix~\ref{app:method-selection}.

\section{Experiments}
\label{sec:experiments}

We investigate the following research questions:

\noindent\textbf{Q1. Overall Effectiveness.}
How does \method compare with existing agents on complex spreadsheet tasks? \textbf{Q2. Grounded Interface.} How does grounded execution affect reliability and correctness of spreadsheet transformations? \textbf{Q3. Policy Learning.} How do different training signals and policy roles influence spreadsheet manipulation performance? \textbf{Q4. Multi-Candidate Consistency.} How can candidate diversity and output agreement improve  solution selection?

\subsection{Experimental Setup}
\label{sec:exp-setup}

\paragraph{Datasets.}
Following Spreadsheet-RL \citep{spreadsheetrl}, we evaluate on two benchmarks. 

\noindent\textbf{\sbench} \citep{spreadsheetbench} contains $912$ complex real-world spreadsheet manipulation tasks with multiple workbook instances per task. \textbf{\dbench} \citep{spreadsheetrl} contains $1,660$ tasks across finance, supply chain, human resources, sales, and real estate.
For training, we sample from $5,925$ real-world Excel tasks released by Spreadsheet-RL~\cite{spreadsheetrl}.

\paragraph{Baselines.}
We compare with following baselines:

\textbf{Closed-source systems.}
We include GPT-4o \citep{gpt4o}, OpenAI o3 \citep{gpto3}, ChatGPT Agent \citep{spreadsheetrl}, Microsoft Copilot Agent Mode \citep{copilot}, and Gemini \citep{gemini} in Google Sheets. For GPT-4o, we report results using LibreOffice and Microsoft Excel.
For ChatGPT Agent, we include both the standard configuration and
the version with additional \texttt{.xlsx} file access.
All results are taken from published evaluations or the official
SpreadsheetBench leaderboard.

\textbf{Open-source models and agents.}
We include Qwen3 models of different sizes \citep{qwen3}. We also compare with TabClaw \citep{tabclaw} and its reported baselines, including SheetAgent \citep{sheetagent}, SpreadsheetLLM \citep{spreadsheetllm}, and ReAct \citep{react}, all evaluated using DeepSeek-V3.2~\cite{deepseek}.

\textbf{Same-backbone comparisons.}
We compare four Qwen3-4B-Thinking-2507 configurations reported by Spreadsheet-RL \citep{spreadsheetrl}: the base model, the model with a spreadsheet-native harness, the model with comprehensive spreadsheet tools, and outcome-based RL trained model. Using the same backbone, we also evaluate our \method framework both with and without training.

\paragraph{Implementation Details.}
We use Qwen3-4B-Thinking-2507 as the backbone model and execute spreadsheet transformations in Python using \texttt{openpyxl}. We train the policies with GRPO and LoRA using \texttt{verl} \citep{verl}
and \texttt{vLLM} \citep{vllm}. For evaluation, we recalculate generated workbooks using Microsoft Excel on Windows and compare them with reference workbooks using the official Spreadsheet-RL comparator.
Training hyperparameters and inference settings are detailed in Appendix~\ref{app:exp-settings}.

\paragraph{Evaluation Metrics.}

We report \textit{Average Accuracy} (same as the Pass@1 results on Spreadsheet-RL) on both datasets, averaging instance accuracy across tasks.
For \sbench, which typically provides three related instances per task, we additionally report \textit{Hard Accuracy}, requiring all instances to be correct. We also report \textit{Execution Success Rate}, measuring the proportion of workbook instances on which the generated program executes successfully. 
All reported accuracy for our models are averaged over three random seeds.

\subsection{Main Results (Q1)}
\label{sec:main-results}


\begin{table}[t]
\centering
\footnotesize
\setlength{\tabcolsep}{3.5pt}
\renewcommand{\arraystretch}{0.94}

\begin{tabular}{@{}p{0.51\columnwidth}rrr@{}}
\toprule
Method &
\multicolumn{3}{c}{Avg. (\%)} \\
\midrule

\rowcolor{black!8}
\multicolumn{4}{c}{\textbf{Closed-source systems}} \\
GPT-4o (LibreOffice) & \multicolumn{3}{c}{16.8} \\
GPT-4o (Excel) & \multicolumn{3}{c}{18.4} \\
OpenAI o3 & \multicolumn{3}{c}{23.3} \\
ChatGPT Agent & \multicolumn{3}{c}{35.3} \\
ChatGPT Agent (.xlsx) & \multicolumn{3}{c}{45.5} \\
Copilot Agent Mode & \multicolumn{3}{c}{57.7} \\
Gemini in Google Sheets & \multicolumn{3}{c}{70.5} \\

\addlinespace[3pt]
\rowcolor{black!8}
\multicolumn{4}{c}{\textbf{Open-source Qwen3 models}} \\
Qwen3-4B-Instruct-2507 & \multicolumn{3}{c}{9.3} \\
Qwen3-4B & \multicolumn{3}{c}{11.0} \\
Qwen3-8B & \multicolumn{3}{c}{15.9} \\
Qwen3-14B & \multicolumn{3}{c}{15.0} \\
Qwen3-32B & \multicolumn{3}{c}{17.6} \\

\addlinespace[3pt]
\midrule
Method & Hard & Avg. & Exec. \\
\midrule

\rowcolor{black!8}
\multicolumn{4}{c}{\textbf{Spreadsheet agents (DeepSeek-V3.2)}} \\
ReAct & 16.1 & 18.8 & \\
SheetAgent & 18.5 & 22.3 & \\
SpreadsheetLLM & 19.8 & 23.3 & \\
TabClaw & 24.5 & 28.8 & \\

\addlinespace[3pt]
\rowcolor{black!8}
\multicolumn{4}{c}{\textbf{Qwen3-4B-Thinking-2507}} \\
\multicolumn{4}{l}{\textit{Spreadsheet-RL}} \\
+ Native harness & & 15.6 & \\
+ Comprehensive tools & & 19.3 & \\
+ Outcome-based RL & & 23.4 & \\
\addlinespace[2pt]
\hdashline
\addlinespace[2pt]

Base model  & 11.4 & 13.5 & 69.1 \\

\addlinespace[2pt]
\hdashline
\addlinespace[2pt]
\multicolumn{4}{l}{\textit{\method} (ours)} \\
+ Grounded task interface  & 23.9 & 29.2 & 97.1 \\
+ Stage-isolated learning & 25.2 & 30.4 & 97.7 \\
\textbf{+ Execution consistency}
& \textbf{27.1} & \textbf{32.8} & \textbf{98.6} \\

\bottomrule
\end{tabular}

\caption{Main results on SpreadsheetBench (\%). Avg. (Average Accuracy) measures the proportion of correctly solved workbook
instances. Hard (Hard Accuracy) requires all instances of a task to be correct. Exec. (Execution Success Rate) measures successful program executions across instances. Execution consistency uses $K=4$ candidates.}

\label{tab:main-spreadsheetbench}
\end{table}


\begin{table}[t]
\centering
\footnotesize
\setlength{\tabcolsep}{2pt}
\renewcommand{\arraystretch}{1.05}

\begin{tabular*}{\columnwidth}{
@{\extracolsep{\fill}}lrrrrr@{}}
\toprule
& \multicolumn{2}{c}{Spreadsheet-RL}
& \multicolumn{3}{c}{\method} \\
\cmidrule(lr){2-3}
\cmidrule(lr){4-6}
Domain & Base & RL & Harness & Learn.
& \shortstack[c]{Consist.} \\
\midrule
Finance-B    & 15.6 & 29.3 & 36.7 & 37.0 & \textbf{42.9} \\
Finance-I    & 7.7  & 16.2 & 26.8 & 28.1 & \textbf{31.2} \\
Finance-A    & 8.1  & 19.3 & 22.2 & 27.4 & \textbf{28.8} \\
Supply Chain & 1.1  & 5.0  & 10.6 & 14.4 & \textbf{16.0} \\
HR           & 0.5  & 3.2  & 13.5 & 14.1 & \textbf{14.6} \\
Sales        & 1.2  & 5.8  & 8.1  & 8.1  & \textbf{11.5} \\
Real Estate  & 1.1  & 1.1  & 10.1 & 11.2 & \textbf{19.1} \\
\midrule
\textbf{Overall}
& 8.4 & 17.2 & 24.9 & 26.2 & \textbf{30.1} \\
\bottomrule
\end{tabular*}

\caption{Average accuracy (\%) on Domain-Spreadsheet using Qwen3-4B-Thinking-2507 as backbone. Harness denotes our grounded task interface, Learn. adds stage-isolated training, and Consist. applies execution consistency with $K=4$ candidate programs.}
\label{tab:main-domain}
\end{table}


\paragraph{SpreadsheetBench.}
Table~\ref{tab:main-spreadsheetbench} compares \method with existing spreadsheet agents. Using Qwen3-4B-Thinking-2507 as its backbone, our grounded task interface improves average accuracy from $13.5\%$ to $29.5\%$ without training, a gain of $16.0$ percentage points. Stage-isolated learning further increases average accuracy to $30.4\%$ and hard accuracy to $25.2\%$.
The resulting average accuracy exceeds the $23.4\%$ reported by Spreadsheet-RL with the same backbone by $7.0$ percentage points.

Execution consistency further improves average accuracy from $30.4\%$ to $32.8\%$ and hard accuracy from $25.2\%$ to $27.1\%$, gains of $2.4$ and $1.9$ percentage points, respectively. Across three random seeds, \method shows a standard deviation of $0.4$ percentage points in average accuracy. Execution success also increases from $97.7\%$ to $98.6\%$. Overall, the grounded task interface provides the largest accuracy gain, with further improvements from policy training and execution consistency.

\paragraph{Domain-Spreadsheet.}
Table~\ref{tab:main-domain} reports results across seven domains. Our grounded task interface (training-free) achieves $24.9\%$ average accuracy, exceeding the $17.2\%$ reported by Spreadsheet-RL. Stage-isolated training improves average accuracy to $26.2\%$. Execution consistency selection raises it to $30.1\%$, a gain of $3.9$
percentage points over \method without consistency.

The consistency setting achieves the highest accuracy among our methods across seven domains, with improvements in Finance-B from $37.0\%$ to $42.9\%$
and real estate from $11.2\%$ to $19.1\%$.
Its execution success rate reaches $99.5\%$, compared with $92.1\%$ for the grounded task interface. Across three random seeds, its average accuracy has a standard deviation of $0.3$ percentage points.

\subsection{Grounded Interface Analysis (Q2)}
\label{sec:grounding-analysis}

We examine the contribution of each component by removing it
from the grounded task interface while keeping other settings fixed.

\begin{table}[t]
\centering
\footnotesize
\setlength{\tabcolsep}{3pt}
\renewcommand{\arraystretch}{1.05}
\begin{tabular}{@{}lrrr@{}}
\toprule
\multicolumn{4}{c}{\textbf{SpreadsheetBench}} \\
\midrule
Configuration & Hard & Avg. & Exec. \\
\midrule
Full interface & \textbf{23.9} & \textbf{29.2} & \textbf{97.1} \\
w/o task-guided inspection & 17.2 & 21.4 & 96.7 \\
w/o API grounding & 20.7 & 25.2 & 95.8 \\
w/o diagnostic repair & 20.2 & 24.4 & 90.5 \\
\midrule
\multicolumn{4}{c}{\textbf{Domain-Spreadsheet}} \\
\midrule
\multicolumn{2}{@{}l}{Configuration} & Avg. & Exec. \\
\midrule
\multicolumn{2}{@{}l}{Full interface}
& \textbf{24.9} & \textbf{92.1} \\
\multicolumn{2}{@{}l}{w/o task-guided inspection}
& 23.3 & 89.6 \\
\multicolumn{2}{@{}l}{w/o API grounding}
& 24.0 & 91.8 \\
\multicolumn{2}{@{}l}{w/o diagnostic repair}
& 20.5 & 82.3 \\
\bottomrule
\end{tabular}
\caption{Component ablations of the grounded task interface on SpreadsheetBench and Domain-Spreadsheet (\%). Each variant removes one component from the full interface. Removing diagnostic repair retains diagnostic checks but
disables LLM-based program correction.}
\label{tab:ablation-grounding}
\end{table}

Table~\ref{tab:ablation-grounding} shows that task-guided inspection has the largest impact on correctness, with its removal reducing average accuracy by $6.7$ percentage points while execution success remains nearly unchanged. Removing API grounding decreases average accuracy by
$3.2$ points. Diagnostic repair improves both correctness and execution reliability, contributing $4.8$ points to average accuracy and $6.6$ points to execution success.

On domain tasks, diagnostic repair has the largest impact, contributing $4.4$ percentage points to average accuracy and $9.8$ points to execution success. Task-guided inspection and API grounding contribute $1.6$ and $0.9$ points to average accuracy, respectively.  Overall, task-guided inspection improves accuracy on SpreadsheetBench, while diagnostic repair enhances accuracy and execution reliability, particularly on Domain-Spreadsheet. Prompt templates are provided in
Appendix~\ref{app:prompt-templates}.

\subsection{Policy Learning Analysis (Q3)}
\label{sec:policy-learning}

We investigate how training signals and stage-isolated
policy learning affect spreadsheet manipulation performance.


\begin{table}[t]
\centering
\small
\setlength{\tabcolsep}{2pt}
\renewcommand{\arraystretch}{1.12}
\begin{tabular*}{\columnwidth}{@{\extracolsep{\fill}}lrrrrr@{}}
\toprule
& \multicolumn{3}{c}{SpreadsheetBench}
& \multicolumn{2}{c}{Domain} \\
\cmidrule(lr){2-4}
\cmidrule(l){5-6}
Configuration & Hard & Avg. & Exec. & Avg. & Exec. \\
\midrule
\multicolumn{6}{@{}l}{\textit{Training signals}} \\
Outcome-based & 24.0 & 29.1 & 95.7 & 22.3 & 90.8\\
Common (multi-signal) & 24.5 & 29.3 & 96.4 & 24.0 & 92.5 \\
\midrule
\multicolumn{6}{@{}l}{\textit{Stage-isolated training}} \\
Generation + Common & 24.9 & 30.0 & 97.0 & 25.3 & 92.7 \\
Generation + Review & \textbf{25.2} & \textbf{30.4}
& \textbf{97.7} & \textbf{26.2} & \textbf{93.2} \\
\bottomrule
\end{tabular*}
\caption{Effects of training signals and stage-isolated
policy learning on SpreadsheetBench and Domain-Spreadsheet
(\%).}
\label{tab:policy-learning}
\end{table}

\paragraph{Training Signals.}
Table~\ref{tab:policy-learning} compares our multi-signal reward in Eq.~\eqref{eq:initial-reward} with an outcome-based reward inspired by Spreadsheet-RL \citep{spreadsheetrl} under the same harness. Our reward combines correctness, execution success, and program quality, providing richer supervision than outcome-based rewards. The common policy increases average accuracy from $29.1\%$ to $29.7\%$ on SpreadsheetBench and from $22.3\%$ to $24.0\%$ on Domain-Spreadsheet, suggesting benefits from richer training signals.

\paragraph{Stage-isolated Training.}
Starting from the common policy, generation and review are separately optimized using stage-isolated edit-progress rewards in Eq.~\eqref{eq:step-reward}. Both policies are optimized using stage-isolated edit-progress rewards. Generation training encourages completion of required workbook edits, while review training targets remaining intent misalignment in executable
programs without disrupting correct edits.
As shown in Table~\ref{tab:policy-learning}, generation specialization increases average accuracy from $29.3\%$ to $30.0\%$ on SpreadsheetBench and from $24.0\%$ to $25.3\%$ on Domain-Spreadsheet, while execution success improves from $96.4\%$ to $97.0\%$ and from $92.5\%$ to $92.7\%$, respectively. Incorporating the review policy raises average accuracy to $30.4\%$ and $26.2\%$, with execution success reaching $97.7\%$ and $93.2\%$, respectively.

Training dynamics are presented in Appendix~\ref{app:training-dynamics}.
Both policies show increasing rewards, with more pronounced fluctuations in generation, while review policy maintains higher execution
success due to its focus on executable programs.

\subsection{Candidate Consistency Analysis (Q4)}
\label{sec:consistency-analysis}

We evaluate candidate composition and selection strategies using four programs ($K=4$) sampled from either the training-free backbone or a mixture of common and generation policies.

\begin{table}[t]
\centering
\footnotesize
\setlength{\tabcolsep}{2.5pt}
\renewcommand{\arraystretch}{1.08}

\begin{tabular*}{\columnwidth}
{@{\extracolsep{\fill}}llrrr@{}}
\toprule
\multicolumn{5}{c}{\textbf{SpreadsheetBench}} \\
\midrule
Candidate pool & Selection & Hard & Avg. & Exec. \\
\midrule
Common + Generation & First-valid & 25.0 & 30.7 & 98.2 \\
Common + Generation & Learned & 25.8 & 31.4 & 98.4 \\
\hdashline
Grounded task interface & Consistency & 26.1 & 31.5 & 98.2 \\
Common policy & Consistency & 26.6 & 32.4 & \textbf{98.6} \\
Generation policy & Consistency & 26.8 & 32.7 & 98.5 \\
Common + Generation & Consistency
& \textbf{27.1} & \textbf{32.8} & \textbf{98.6} \\
\midrule
\multicolumn{5}{c}{\textbf{Domain-Spreadsheet}} \\
\midrule
Candidate pool & Selection
& \multicolumn{2}{c}{Avg.} & Exec. \\
\midrule
Common + Generation & First-valid
& \multicolumn{2}{c}{24.8} & 99.2 \\
Common + Generation & Learned
& \multicolumn{2}{c}{25.3} & 99.3 \\
\hdashline
Grounded task interface & Consistency
& \multicolumn{2}{c}{27.2} & 99.3 \\
Common policy & Consistency
& \multicolumn{2}{c}{28.4} & 99.2 \\
Generation policy & Consistency
& \multicolumn{2}{c}{29.0} & 99.4 \\
Common + Generation & Consistency
& \multicolumn{2}{c}{\textbf{30.1}} & \textbf{99.5} \\
\bottomrule
\end{tabular*}

\caption{Selection results on SpreadsheetBench and Domain-Spreadsheet (\%) with $K=4$. Common + Generation samples two candidates per policy. The selected program is reviewed once
for intent misalignment.}

\label{tab:consistency-analysis}
\end{table}

\paragraph{Selection Strategies.}

The grounded task interface samples candidates from the original backbone without additional policy training. We compare execution consistency with two alternatives. First-valid selects the first executable candidate, while the learned selector uses Qwen3-4B with a trained LoRA adapter to rank candidates by predicted correctness.
More details are in Appendix~\ref{app:learned-selector}.

As shown in Table~\ref{tab:consistency-analysis},
multi-candidate consistency outperforms both first-valid and learned selection under the same candidate pool. On SpreadsheetBench, it achieves $27.1\%$ hard accuracy and $32.8\%$ average accuracy, improving over first-valid by
$2.1$ percentage points on both metrics and over learned selection by $1.3$ and $1.4$ points, respectively. On Domain-Spreadsheet, average accuracy reaches $30.1\%$, compared with $24.8\%$ for first-valid and $25.3\%$ for learned selection.

\paragraph{Candidate Composition.}

Under consistency selection, mixing policies achieves the highest accuracy on both benchmarks. On SpreadsheetBench, the mixed pool reaches $27.1\%$ hard accuracy, exceeding the common-only ($26.6\%$) and generation-only ($26.8\%$) pools. On Domain-Spreadsheet, it achieves $30.1\%$ average accuracy, compared with $28.4\%$ and $29.0\%$, respectively. These results highlight the benefits of combining complementary policies over single-policy sampling.

\begin{table}[t]
\centering
\footnotesize
\setlength{\tabcolsep}{6pt}
\renewcommand{\arraystretch}{1.08}
\begin{tabular*}{\columnwidth}
{@{\extracolsep{\fill}}lrrr@{}}
\toprule
Benchmark & Selected & Oracle & Gap \\
\midrule
SpreadsheetBench & 32.8 & \textbf{39.7} & 6.9 \\
Domain-Spreadsheet & 30.1 & \textbf{40.7} & 10.6 \\
\bottomrule
\end{tabular*}
\caption{Oracle bounds for Common + Generation ($K=4$) in average accuracy (\%). Oracle selects one program per task using references. Gap measures the difference from consistency selection in percentage points.}
\label{tab:consistency-oracle}
\end{table}

\paragraph{Oracle Upper Bound.}
Table~\ref{tab:consistency-oracle} compares multi-candidate execution consistency with oracle selection under the same candidate pool. On SpreadsheetBench, the oracle reaches $39.7\%$ average accuracy, leaving a $6.9$ point gap over consistency selection. The gap is larger on Domain-Spreadsheet ($10.6$ points), indicating that further improvements in candidate selection remain possible.

\section{Conclusion}

We presented \method, a spreadsheet manipulation framework combining grounded task interaction, stage-isolated policy learning, and execution consistency. On SpreadsheetBench, our grounded task interface improves Qwen3-4B's average accuracy from $13.5\%$ to $29.2\%$ without training. Stage-isolated learning further increases accuracy to $30.4\%$, while execution consistency raises it to $32.8\%$. On Domain-Spreadsheet, \method achieves $30.1\%$ average accuracy, compared with $17.2\%$ for the outcome-based RL baseline. These results demonstrate the effectiveness of combining task grounding, stage-isolated policy learning, and consistency selection for spreadsheet agents.

\newpage
\section*{Limitations}
Our study focuses on spreadsheet manipulation using compact open-weight LLMs with Python and \texttt{openpyxl}. Tasks involving specialized Excel features or external integrations may require additional tools. Task-guided inspection uses bounded workbook observations, which may need further adaptation for exceptionally large or complex workbooks. Moreover, execution consistency introduces additional inference costs by generating multiple candidates, creating a trade-off between computational efficiency and solution reliability.

\section*{Ethics Statement}
All datasets and models used in this paper are publicly available, and their usage strictly adheres to the corresponding licenses and terms of use.


\bibliography{custom}

\newpage
\appendix

\section{Method Details}
\label{app:method-details}
\subsection{Grounding and Execution}
\label{app:method-interface}

\paragraph{Structured task description.}
Before code generation, the model receives the instruction, target locations,
and a compact workbook preview. It returns six fields:
\texttt{where}, \texttt{from}, \texttt{what}, \texttt{write\_mode},
\texttt{risks}, and \texttt{api\_needs}. These fields specify where to edit,
what workbook information to inspect, what operation is requested, how the
result should be written, and any relevant risks or API requirements.

For example, consider a task that asks to fix a circular formula in
\texttt{B323}. The workbook contains

\begin{lstlisting}
B320 = 605
B323 = =IF(A323="E","",MAX(B$7:B377)+1)
\end{lstlisting}

where the range \texttt{B7:B377} includes \texttt{B323} itself. A corresponding
task description can be:

\begin{lstlisting}
{
  "where": "B323",
  "from": "A323 and previous values in column B",
  "what": "fix the circular MAX formula",
  "write_mode": "formulas",
  "risks": ["circular reference"],
  "api_needs": ["cell access", "formula assignment"]
}
\end{lstlisting}

The harness then exposes only the workbook context needed for the task, such as
the target cell, nearby values, relevant formulas, headers, and merged regions.
This grounds generation in the current workbook state.

\paragraph{Observation construction.}
The harness builds a compact view of the task-relevant workbook region, including nearby cells, formulas, headers, merged ranges, and cell types. Formula expressions are preserved during inspection so that the generator can reason over the actual workbook logic.

\paragraph{API lookup.}
The \texttt{api\_needs} field is used to retrieve a small set of relevant \texttt{openpyxl} references, which are provided to the generator together with the workbook observations. The lookup is deterministic and based on task-specific API keywords.

\subsection{Diagnostic repair.}
\label{app:diagnositic_repair}
The harness executes each generated program and checks for clear implementation
failures. When the program fails to run or produces an obviously invalid
workbook, the harness starts a diagnostic repair. For each repair, the model receives the current program, a concise diagnostic, the user instruction, and the relevant workbook and API context. The model
returns a complete corrected program. The harness then reruns it from the
original workbook. Repair stops when no triggering problem remains or after two repair attempts.

Diagnostic repair uses only problems observed during program execution and workbook checking. This mechanism provides a
fast correction path for implementation failures. It complements the intent misalignment review policy introduced in Section~\ref{sec:learning}, which targets programs
that execute successfully but may still fail to reflect the user's intent.

\paragraph{Repair prompt.}
The repair prompt is organized around the current failure and the information
needed to correct it:

\begin{quote}
\small

\textbf{Goal.}  
Repair the current program using the diagnostic. Correct the identified issue, preserve unrelated logic, and follow the same spreadsheet-editing constraints used during generation.

\vspace{4pt}

\textbf{Context.}  
The prompt provides the user instruction, target locations, current program, diagnostic message, focused workbook observations, and relevant
\texttt{openpyxl} references.

\vspace{4pt}

\textbf{Expected output.}  
Return one complete corrected Python program that can be executed directly by the harness.

\end{quote}

\subsection{Policy Optimization and Training Settings}
\label{app:method-training}

\paragraph{GRPO optimization.}
For each training prompt $x$, group relative policy optimization
(GRPO) ~\cite{deepseekmath} samples $K$ responses
$\{y_i\}_{i=1}^{K}$ and assigns each response a reward $r_i$.
We compute the group-relative advantage as

\begin{equation}
A_i =
\frac{r_i-\bar r}{\sigma_r},
\end{equation}

where $\bar r$ and $\sigma_r$ are the mean and standard deviation of rewards
for responses sampled from the same prompt.

For token $t$ in response $y_i$, the policy ratio is

\begin{equation}
\rho_{i,t}(\theta)
=
\frac{
\pi_{\theta}(y_{i,t}\mid x,y_{i,<t})
}{
\pi_{\mathrm{old}}(y_{i,t}\mid x,y_{i,<t})
},
\end{equation}

where $\pi_{\mathrm{old}}$ denotes the policy used to generate the training
responses. We define the clipped ratio

\begin{equation}
\tilde{\rho}_{i,t}
=
\operatorname{clip}
\left(
\rho_{i,t},
1-\epsilon_c,
1+\epsilon_c
\right),
\end{equation}

and optimize

{\small
\begin{equation}
\mathcal{J}_{\mathrm{GRPO}}(\theta)
=
\frac{1}{K}
\sum_{i=1}^{K}
\frac{1}{|y_i|}
\sum_t
\min
\left(
\rho_{i,t}(\theta)A_i,\,
\tilde{\rho}_{i,t}A_i
\right).
\end{equation}
}

The response-level advantage $A_i$ is shared across tokens in the same
response. Our implementation does not add a separate KL regularization term.

\paragraph{LoRA.}
We use low-rank adaptation (LoRA) ~\cite{lora} for
parameter-efficient training. Each trained policy uses its own LoRA adapter, while the backbone parameters remain fixed. Generation policy and review policy training start from the same trained policy and proceed independently.

\paragraph{Training settings.}
Table~\ref{tab:method-training} reports the shared optimization settings
for generation policy and review policy training. Stage-specific data construction is
described below.

\begin{table}[t]
\centering
\small
\begin{tabular}{lr}
\toprule
Setting & Value \\
\midrule
Optimizer updates & 40 \\
Prompts per update & 4 \\
Responses per prompt ($K$) & 4 \\
Learning rate & $8\times10^{-7}$ \\
LoRA rank & 16 \\
LoRA scaling parameter & 32 \\
LoRA dropout & 0 \\
Prompt token limit & 12,288 \\
Response token limit & 12,288 \\
Sampling temperature & 0.8 \\
Top-$p$ & 0.95 \\
Checkpoint interval & 10 updates \\
\bottomrule
\end{tabular}
\caption{Optimization settings for generation policy and review policy training.}
\label{tab:method-training}
\end{table}

\paragraph{Initial training data.}
The common starting policy is trained from a pool of 1,200 spreadsheet tasks. We use $1,136$ tasks for training and 64 for validation. After filtering prompts that exceed the context limit, 1,127 training tasks remain eligible for optimization.

\paragraph{Generation policy curriculum.}
Generation policy training starts from a 256-task candidate pool. We prioritize tasks whose sampled programs contain both a correct solution and an executable but incorrect one. These tasks expose cases where the Generation policy can solve the problem but does not yet do so consistently. We interleave them with training examples from different spreadsheet operation types to maintain broader coverage. Repeated sampling intentionally concentrates training on these unstable cases.

\paragraph{Review policy training.}
Review policy training uses a 256-task pool with multiple executable programs per task. We include both correct and incorrect programs so that the review policy learns when to keep an existing program and when to revise it. Using multiple programs also exposes the review policy to different candidate solutions for the same task.

\subsection{Reward Definitions}
\label{app:method-reward}

We use two reward designs. The first trains the common starting policy.
The second adds edit-level progress for subsequent generation policy and review policy
training.

\paragraph{Initial diagnostic reward.}
A task can be evaluated on multiple workbook instances. Let $S$ denote the
average correctness across these instances, and let $H$ indicate whether all
instances are correct. We summarize task correctness as

\begin{equation}
R_{\mathrm{task}}
=
\eta S + (1-\eta)H,
\end{equation}

where $\eta=0.75$, placing more weight on correctness across individual
workbooks while retaining a signal for complete task success.

The initial reward uses $R_{\mathrm{task}}$ as its main signal, scaled by
$0.82$, with smaller contributions from execution success ($0.10$) and output quality ($0.06$). Executable but incorrect programs receive a small positive reward, while execution failures receive negative penalties ranging from $-0.06$ to $-0.12$ depending on the error type. This design prioritizes task correctness while still encouraging executable and well-formed programs.

\paragraph{Progress reward details.}
The implementation explicitly handles cases with no required or predicted edits
to avoid undefined precision or recall. Reference workbooks in training data are used only for reward computation and evaluation and are not provided to the policy.

\subsection{Execution Consistency Details}
\label{app:method-selection}

\paragraph{Candidate configuration.}
We use four candidates in the fixed order
$[D_0,D_1,S_0,S_1]$. The two $D$ candidates use the common starting policy for generation, while the two $S$ candidates use the separately trained generation policy. Each candidate independently executes the full harness.

\begin{figure*}[t]
    \centering
    \includegraphics[width=0.82\textwidth]{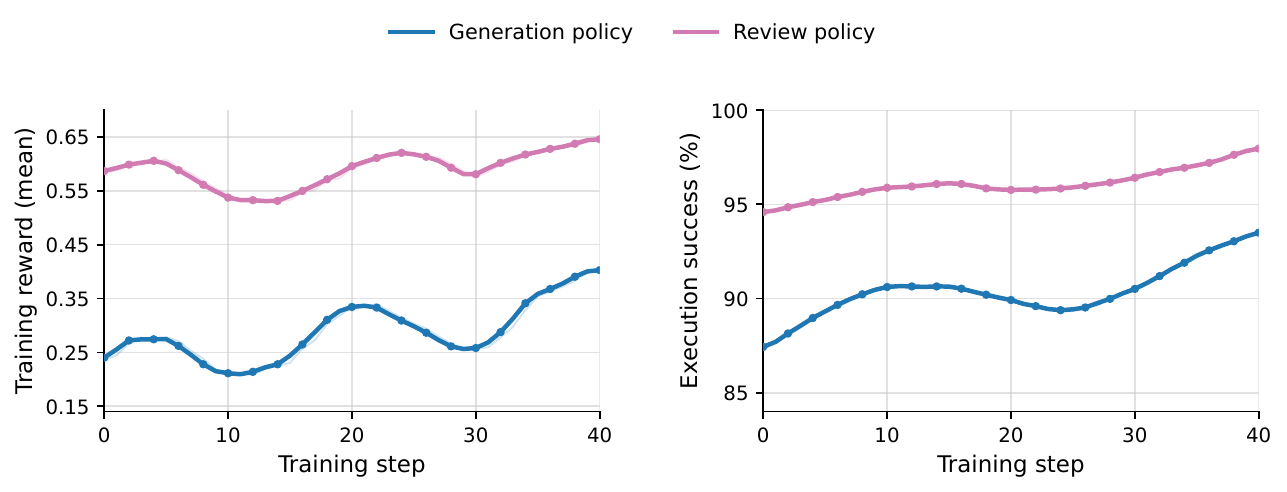}
    \caption{Training dynamics of generation and review policies.
    The panels show mean training reward and execution success
    across optimization steps.}
    \label{fig:training-dynamics}
\end{figure*}

\paragraph{Workbook fingerprints.}
Fingerprints are constructed from the stored contents of each output workbook. They record worksheet names and order, nonempty cell locations, types, values, and formula expressions, together with formatting and layout information such as merged cells, styles, and filters. Formulas are compared as stored expressions without recalculation.

Fingerprinting is limited to $200,000$ inspected cells and $64$ MiB of decompressed workbook content. Outputs that cannot be fingerprinted are excluded from agreement voting.

\paragraph{Selection procedure.}
Candidates are grouped by matching fingerprints. We select from the largest group containing at least two candidates, breaking ties by the fixed candidate order. If no group meets this condition, we choose the first candidate with an inspectable output, or the first candidate if none is available.  Selection also requires no additional review policy calls or learned selection model.

\section{Additional Experimental Details}
\label{app:exp-details}

\subsection{Training and Inference Settings}
\label{app:exp-settings}

Training hyperparameters are provided in
Table~\ref{tab:method-training}. For benchmark inference, we use a sampling temperature of $0.2$ and a context window of $32,768$ tokens. Workbook understanding is limited to $1,024$ output tokens, and program
execution has a timeout of $300$ seconds. Generation and repair outputs are constrained by the remaining context budget.

Each candidate is executed independently on the original input workbook, with at most two diagnostic repair attempts.

\subsection{Prompt Templates}
\label{app:prompt-templates}

We present prompt templates used by the grounded task interface and policy learning stages. Placeholders are populated by the harness at runtime.

\paragraph{Shared System Instruction.}
All stages share the following system instruction.

\begin{lstlisting}[style=prompt]
You write compact, executable Python for spreadsheet manipulation. Use openpyxl only. Programs must generalize across workbook instances of the same task.
\end{lstlisting}

\paragraph{Task-guided Workbook Inspection.}
The model first produces a structured task description to guide focused workbook inspection. The harness then uses this description to retrieve relevant workbook observations.

\begin{lstlisting}[style=prompt]
Analyze the spreadsheet task and return JSON only.
Do not generate code or reasoning traces.

Instruction: {instruction}
Instruction type: {instruction_type}
Target locations: {answer_position}
Workbook preview: {workbook_preview}

Return exactly:
{
  "where": "target sheets and ranges",
  "from": "source data to inspect",
  "what": "required operations",
  "write_mode": "static_values | formulas |
                 formatting | structure | unknown",
  "risks": [],
  "api_needs": []
}

Include only task-relevant details.
Use empty lists when uncertain.
\end{lstlisting}

\paragraph{API Grounding.}
Based on the identified API requirements, the harness retrieves relevant \texttt{openpyxl} signatures, usage notes, and documentation snippets. These references are inserted into generation and diagnostic repair prompts.

\begin{lstlisting}[style=prompt]
Relevant openpyxl API references:

{api_name}{installed_signature}
Usage: {usage_note}
Documentation: {short_installed_docstring}

Canonical examples:
ws.cell(row=r, column=c).value = value
ws.delete_rows(idx=row_index, amount=1)
for row in ws.iter_rows(min_row=2):
    values = [cell.value for cell in row]
\end{lstlisting}

\paragraph{Program Generation.}
The generation policy receives the task description, focused workbook observations, and API references to produce an executable transformation program.

\begin{lstlisting}[style=prompt]
Generate Python/openpyxl code for the requested
workbook transformation.

Return one Python code block defining:
    transform(wb) -> None

Requirements:
- Do not load, save, or access workbook files.
- Use openpyxl and the Python standard library.
- Apply edits to the intended sheets and ranges.
- Preserve unrelated cells, formulas, and styles.
- Prefer static values for computed outputs;
  use formulas when explicitly requested.
- Handle missing values, numeric strings, and
  merged cells conservatively.
- Avoid unsupported openpyxl APIs.
- Generalize across instances of the same task.

Mode: {mode}
Instruction type: {instruction_type}
Target locations: {answer_position}
Instruction: {instruction}
Workbook preview: {workbook_preview}
Task description: {understanding_json}
Focused inspection: {focused_inspection}
API references: {api_references}
\end{lstlisting}

\paragraph{Execution and Diagnostic Repair.}
When execution or output checks detect a problem,
the harness provides a concise diagnostic to the
generation policy. The revised program is rerun from the original workbook, with at most two repair attempts.

\begin{lstlisting}[style=prompt]
Repair the current transform(wb) program using
the supplied diagnostic.

- Fix the identified issue without replanning.
- Preserve correct edits and unrelated logic.
- Address execution failures, invalid outputs,
  or verification errors.
- For value outputs, avoid uncached formulas.
- Return one Python code block containing
  the complete repaired transform(wb).
- Do not load or save workbooks.

Instruction: {instruction}
Target locations: {answer_position}
Diagnostic: {diagnostic_json}

Current program:
{current_code}

Task description: {understanding_json}
Workbook context: {focused_workbook_context}
API references: {api_references}
\end{lstlisting}

\paragraph{Review Policy.}
After candidate selection, the review policy performs one pass to check the selected program for intent misalignment. It may retain or revise the program based on the available input and output observations.

\begin{lstlisting}[style=prompt]
Review the program against the user's instruction.
Successful execution does not guarantee correctness.

Use only the supplied workbook observations.
Incomplete observations are not evidence of errors.

Return one of:
DECISION: KEEP
- Provide a brief supporting reason.

DECISION: REVISE
- Explain the identified issue briefly.
- Return the complete corrected transform(wb).

Preserve valid edits and unrelated workbook content.
Do not load, save, or access files.

Instruction: {instruction}

Current program:
{current_code}

Input/output observations:
{bounded_input_output_observations_json}
\end{lstlisting}

\subsection{Excel Recalculation and Evaluation}
\label{app:excel-evaluation}

\paragraph{Recalculation.}
Since \texttt{openpyxl} does not evaluate Excel formulas, we use Microsoft Excel on Windows to recalculate generated workbooks before final evaluation. The resulting workbooks are compared with reference workbooks using the official Spreadsheet-RL comparator.

\paragraph{Workbook comparison.}
The comparator evaluates the specified target regions after normalizing cell values. Numbers and numerical strings are rounded to two decimal places, dates are converted to Excel serial values, and times are
normalized to HH:MM. Empty strings and missing values are treated as equivalent. Formula cells are compared using their recalculated values. A workbook instance is considered correct only if all target cells match the reference.

\paragraph{Recalculation scope.}
Excel recalculation is used only for final evaluation and does not affect policy training, diagnostic repair, or candidate selection.

\subsection{Policy Training Dynamics}
\label{app:training-dynamics}

We examine the training dynamics of the generation and review policies in terms of mean training reward and execution success. The two policies optimize different stages of spreadsheet
manipulation: generation constructs workbook transformations, while review addresses intent misalignment in executable programs.

\subsection{Learned Selector}
\label{app:learned-selector}

We train a lightweight candidate selector using
Qwen3-4B-Thinking-2507 with LoRA. For each candidate program,
the selector receives the user instruction, candidate code,
and bounded observations of the input and output workbooks.

\paragraph{Training Labels.}
Candidate correctness is determined using the reference workbooks during training.  For each training task, a candidate is labeled $\texttt{1}$ only if its resulting workbooks are correct
across all available instances, and $\texttt{0}$ otherwise. Executable programs with incorrect outputs are also labeled $\texttt{0}$.
Samples with missing outputs, invalid reference comparisons, or observations exceeding the input budget are excluded. Reference workbooks are used only for label construction and are never included in selector inputs.

\paragraph{Training and Selection.}
We freeze the backbone model and optimize only the LoRA
parameters using cross-entropy loss over the two label
tokens, $\texttt{0}$ and $\texttt{1}$. At inference,
each candidate is scored by the normalized probability

\begin{equation}
s(c) =
\frac{\exp(\ell_1(c))}
{\exp(\ell_0(c))+\exp(\ell_1(c))},
\end{equation}

where $\ell_0(c)$ and $\ell_1(c)$ denote the logits
of the two label tokens for candidate $c$.
The selector chooses the candidate with the highest
score, without generating additional reasoning or code.

\end{document}